\documentclass[letterpaper, 10 pt, conference]{ieeeconf}  

\IEEEoverridecommandlockouts                              

\usepackage{amsmath} 
\usepackage{amssymb}  
\usepackage{graphicx}
\usepackage{booktabs}
\usepackage[table]{xcolor} 
\usepackage{multirow}       
\usepackage[skip=3pt, font={small}]{caption}
\usepackage[colorlinks=false]{hyperref}
\usepackage{bm}
\definecolor{lightgreen}{RGB}{220, 237, 191}
\definecolor{lightblue}{RGB}{221, 235, 247}
\definecolor{firebrick}{HTML}{B22222}   

\renewcommand{\baselinestretch}{1.1} 
\normalsize
\newcommand{\method}{VMS}

\title{\LARGE \bf
\textsc{KungfuBot}{\textbf{\textcolor{firebrick}{2}}}: Learning Versatile Motion Skills for \\
Humanoid Whole-Body Control 
}

\author{Jinrui Han$^{1,2}$ \quad Weiji Xie$^{1,2}$ \quad Jiakun Zheng$^{1,3}$ \quad Jiyuan Shi$^{1}$\quad Weinan Zhang$^{2}$ \quad Ting Xiao$^{3}$ \quad Chenjia Bai$^{\dag 1}$
\thanks{$\dag$Corresponding Author}
\thanks{$^{1}$Institute of Artificial Intelligence (TeleAI), China Telecom, $^{2}$Shanghai Jiao Tong University, $^{3}$East China University of Science and Technology}
}

\makeatletter
\let\@oldmaketitle\@maketitle
\renewcommand{\@maketitle}{\@oldmaketitle
  \centering
  \setcounter{figure}{0}%
  \begin{minipage}{\linewidth}
    \includegraphics[width=\textwidth]{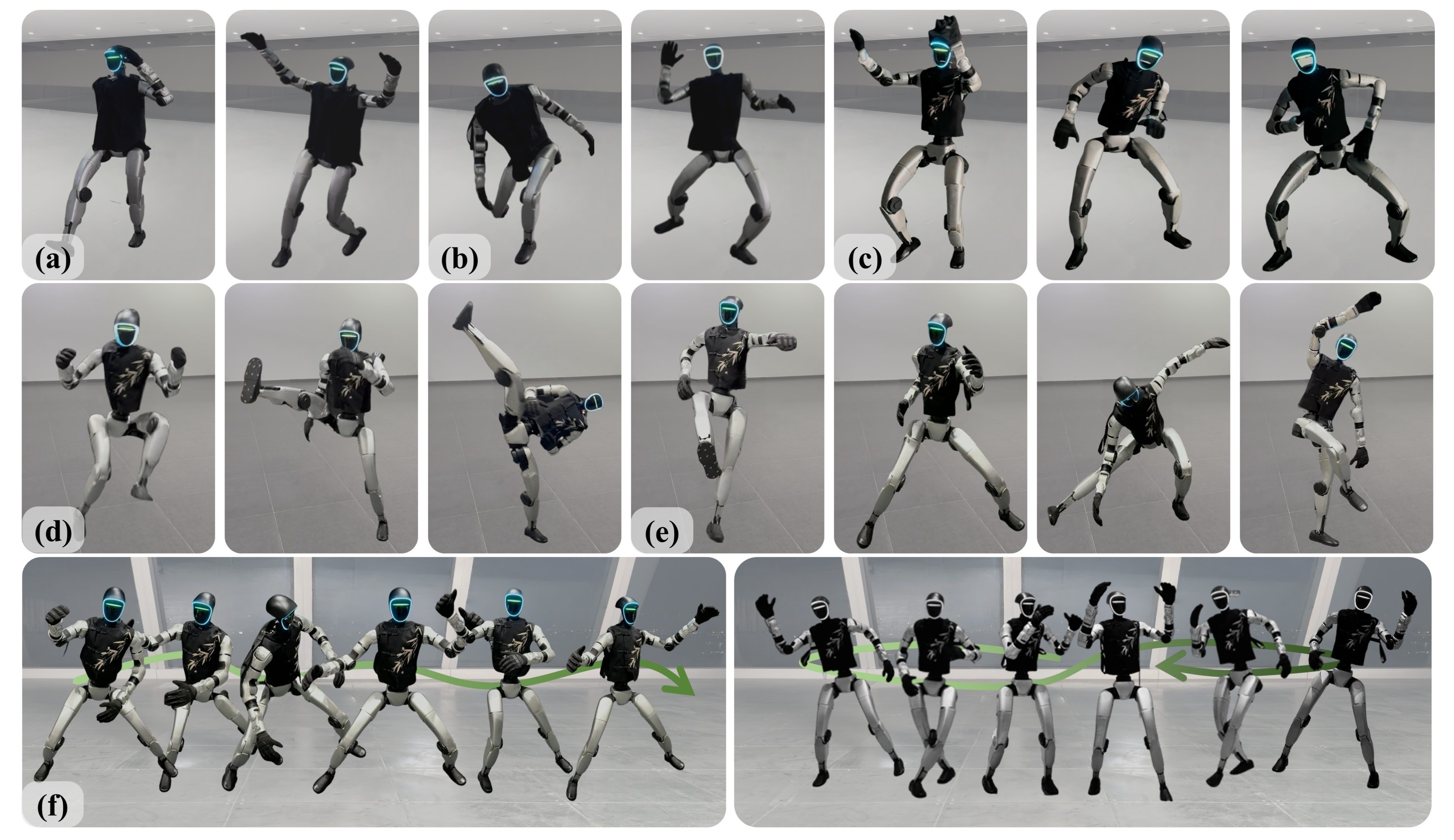}
    {\captionsetup{hypcap=false}%
    \captionof{figure}{\label{fig:cover}\textbf{Humanoid learning versatile motion skills.}
    We deploy VMS on the Unitree G1 humanoid robot, demonstrating its capability to perform a broad category of motion skills with strong stability and generalization. The repertoire includes (a) walking and running, (b) ball throwing and racket swinging, (c) dancing, (d) diverse kicking, (e) Kung Fu and (f) long sequences of martial arts and dance.}}
  \end{minipage}
  \vspace{0pt}
}
\makeatother

\begin{document}

\maketitle
\thispagestyle{empty}
\pagestyle{empty}


\begin{abstract}

Learning versatile whole-body skills by tracking various human motions is a fundamental step toward general-purpose humanoid robots. This task is particularly challenging because a single policy must master a broad repertoire of motion skills while ensuring stability over long-horizon sequences. To this end, we present \method{}, a unified whole-body controller that enables humanoid robots to learn diverse and dynamic behaviors within a single policy. Our framework integrates a hybrid tracking objective that balances local motion fidelity with global trajectory consistency, and an Orthogonal Mixture-of-Experts (OMoE) architecture that encourages skill specialization while enhancing generalization across motions. A segment-level tracking reward is further introduced to relax rigid step-wise matching, enhancing robustness when handling global displacements and transient inaccuracies. We validate \method{} extensively in both simulation and real-world experiments, demonstrating accurate imitation of dynamic skills, stable performance over minute-long sequences, and strong generalization to unseen motions. These results highlight the potential of \method{} as a scalable foundation for versatile humanoid whole-body control. The project page is available at \href{https://kungfubot2-humanoid.github.io}{Kungfubot2-humanoid.github.io}.

\end{abstract}

\section{Introduction}
Humanoid robots hold great potential for imitating human behaviors, spanning stable locomotion to agile, complex motions. Realizing this capability requires a universal whole-body controller that can generalize across versatile skills. Recent advances in motion capture (Mocap) systems have enabled the collection of large-scale human motion datasets, offering rich resource for developing such controllers.

Existing research~\cite{2022-TOG-ASE,phc,tessler2024maskedmimic,yu2025skillmimic} in physics-based character animation has explored learning-based methods for building controllers that mimic versatile, human-like behaviors in simulation. Motivated by these advances, recent work has extended this paradigm to humanoid robots~\cite{OmniH2O,humanplus,expressive,serifi2024vmp,he2025hover}, addressing challenges such as partial observability~\cite{OmniH2O,exbody2}, physical plausibility~\cite{xie2025kungfubotphysicsbasedhumanoidwholebody}, and the sim-to-real gap~\cite{he2025asapaligningsimulationrealworld}. However, achieving high-fidelity imitation often requires training separate policies for each motion~\cite{xie2025kungfubotphysicsbasedhumanoidwholebody,he2025asapaligningsimulationrealworld,liao2025beyondmimicmotiontrackingversatile}, which limits generalization. Several methods attempt to learn a single policy for multiple motions, but these approaches are constrained by limited policy expressiveness, typically relying on a single MLP network~\cite{ze2025twist,yin2025unitrackerlearninguniversalwholebody}, and lack mechanisms to balance local and global tracking objectives~\cite{li2025clone,chen2025gmtgeneralmotiontracking}. Specifically, local tracking (e.g., velocities or relative keybody poses) can reduce error accumulation but may compromise global stability~\cite{chen2025gmtgeneralmotiontracking}, whereas global tracking ensures overall coherence but is prone to long-horizon drift~\cite{xie2025kungfubotphysicsbasedhumanoidwholebody}. Thus, two central challenges remain: enhancing policy expressiveness for versatile skill learning, and reconciling local and global tracking to achieve both motion fidelity and long-horizon stability.

To this end, we present \method{}, a universal whole-body controller that enables humanoid robots to learn versatile motion skills. \method{} first introduces a hybrid tracking objective that preserves local motion pose while mitigating global drift, addressing the instability of purely local or global tracking. Second, an Orthogonal Mixture-of-Experts (OMoE) architecture disentangles skill representations, improving policy expressiveness and reducing overlap between skill representations. Third, a segment-level tracking reward further improves robustness by relaxing rigid step-wise matching, enabling stable long-horizon motion execution. Extensive experiments demonstrate that \method{} performs dynamic skills with high fidelity and sustains stable tracking over minute-level sequences. Our main contributions are:

\begin{itemize}
 \item We propose an OMoE architecture that disentangles motion representations, improving policy expressiveness and generalization across diverse skills.
 \item We introduce a hybrid tracking objective together with a segment-level reward, balancing motion style fidelity and long-horizon stability.
 \item We validate \method{} on extensive simulated and real-world experiments, achieving robust tracking of both high-dynamic and minute-level motion sequences.
\end{itemize}
\section{Related Work}
\subsection{Humanoid Whole-Body Control}
Traditional model-based approaches to humanoid whole-body control rely on accurate dynamics models for precise task execution~\cite{geyer2003positive,sreenath2011compliant}, but they demand complex modeling effort and generalize poorly across diverse skills or unmodeled dynamics.
Learning-based approaches, in contrast, typically depend on manually crafted, task-specific rewards. While such methods have been successfully applied to locomotion on challenging terrain~\cite{wang2024beamdojo,xie2025humanoid,wang2025moremixtureresidualexperts}, jumping~\cite{li2023robust}, parkour~\cite{zhuang2025humanoid}, and fall recovery~\cite{huang2025learning,he2025learning}, each task demands extensive reward engineering, and generating human-like motions remains difficult~\cite{peng2021amp}. To handle the distinct objectives of upper and lower body control, some works decompose the solution into separate policies~\cite{zhang2025falconlearningforceadaptivehumanoid,li2025hold,almi}, though this often limits coordination and generalization. Others tackle complex tasks such as table tennis via hierarchical planning and learning~\cite{su2025hitterhumanoidtabletennis}.

In contrast, whole-body motion tracking directly leverages human motion data as a reference, providing a unified control objective for the entire body: reproducing the reference motion. This eliminates the need for task-specific reward design and naturally encourages human-like coordination and expressiveness across a wide range of skills.

\subsection{Humanoid Motion Tracking}
Humanoid motion tracking aims to learn lifelike behaviors directly from human motion data. DeepMimic~\cite{deepmimic} pioneers a phase-based tracking framework that combines random state initialization with early termination to imitate individual motions. To bridge the sim-to-real gap, ASAP~\cite{he2025asapaligningsimulationrealworld} proposes a multistage training pipeline with a delta-action model for dynamic skills. HuB~\cite{zhang2025hub} and KungfuBot~\cite{xie2025kungfubotphysicsbasedhumanoidwholebody} employ elaborate motion processing and tracking mechanisms, achieving accurate imitation of highly dynamic single motions.

For learning diverse humanoid motions within a single policy, OmniH2O~\cite{OmniH2O} introduces a universal controller that inspired subsequent humanoid works. ExBody2~\cite{exbody2} improves expressiveness by decomposing tracking targets and applying motion filtering. TWIST~\cite{ze2025twist} and CLONE~\cite{li2025clone} enable high-quality tracking but are tailored to teleoperation settings and focus on relatively low-dynamic motions. BumbleBee~\cite{wang2025expertsgeneralistgeneralwholebody} adopts a two-stage strategy: clustering motions and training separate experts, followed by policy distillation. GMT~\cite{chen2025gmtgeneralmotiontracking} achieves robust tracking of highly dynamic motions by emphasizing root velocity and pose over global positions. UniTracker~\cite{yin2025unitrackerlearninguniversalwholebody} supports dynamic movements, though its reliance on global targets limits stability in executing long-sequence motions. More recently, BeyondMimic~\cite{liao2025beyondmimicmotiontrackingversatile} demonstrates high-fidelity tracking of single motions through well-designed objectives and precise system identification, and uses a distilled unified diffusion policy for task-specific control.
Building on these advances, our work focuses on learning a single universal policy capable of reproducing diverse, long-sequence motions with both local style fidelity and global stability.
\section{Method}
\begin{figure*}
    \centering
    \includegraphics[width=\linewidth]{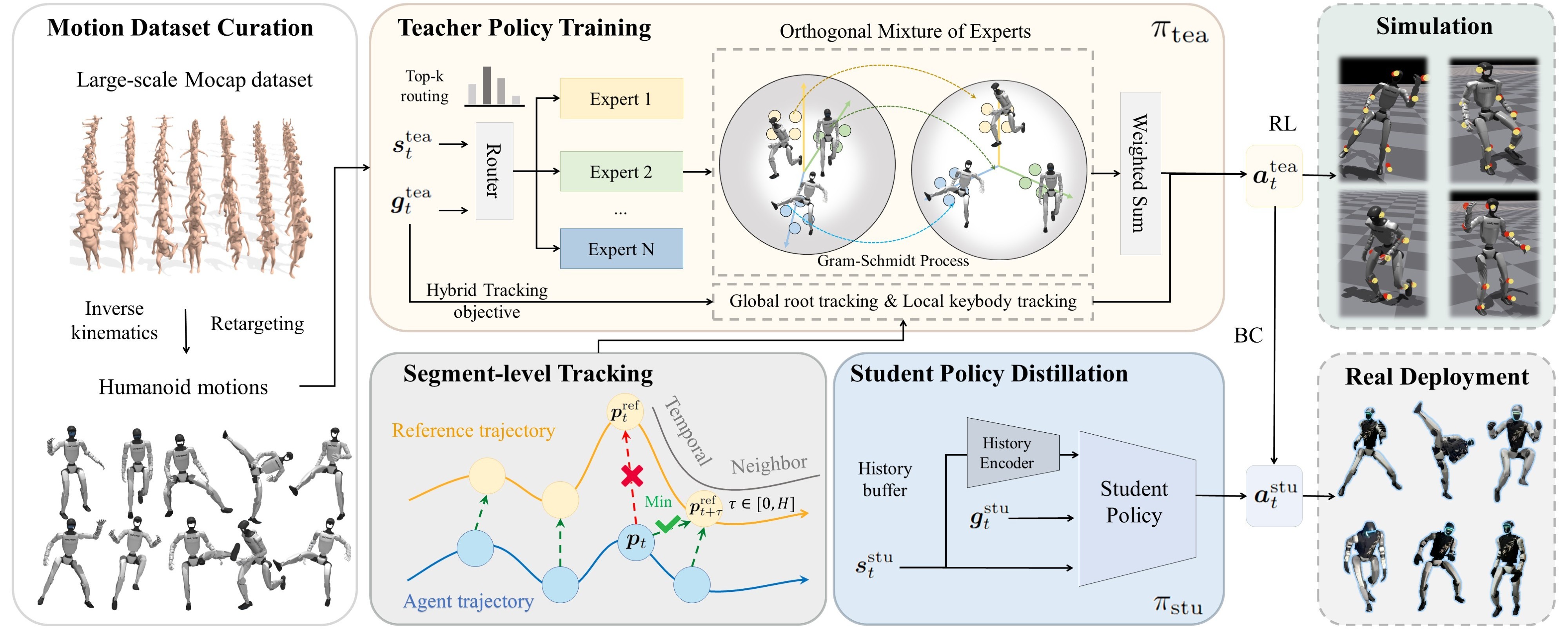}
    \caption{\textbf{Framework of \method{}.} 
The large-scale motion capture dataset is first retargeted to the humanoid skeleton using an IK-based method. A teacher policy $\pi_{\mathrm{tea}}$ is then trained in simulation with a hybrid tracking objective, further enhanced by a segment-level reward for long-horizon robustness. The student policy $\pi_{\mathrm{stu}}$ is distilled from the teacher policy through behavior cloning. Finally, the distilled policy is deployed on the real humanoid robot.}
    \label{fig:framework}
\end{figure*}
\subsection{Problem Definition} 
In this work, we adopt the Unitree G1 robot~\cite{unitree-g1}, controlling 23 degrees of freedom (DoF), excluding the three DoFs of each wrist. We formulate humanoid motion tracking as a goal-conditioned reinforcement learning (RL) problem, where the agent interacts with the environment according to a policy $\pi$ to maximize cumulative reward. At each timestep $t$, the policy receives the state $\bm s_t$ and target state $\bm g_t$, and outputs an action $\bm a_t \in \mathbb{R}^{23}$, which is subsequently converted into motor torques via a PD controller. This defines the policy as $\pi(\bm a_t | \bm s_t, \bm g_t)$. The environment dynamics $p(\bm s_{t+1} | \bm s_t, \bm a_t)$ determine the next state, and a dense reward $r( \bm s_t, \bm g_t, \bm a_t)$ evaluates tracking performance while providing regularization. The agent's objective is to maximize the expected discounted return $J = \mathbb{E}\Big[\sum_{t=0}^{T-1} \gamma^t r_t \Big]$.

Following prior work~\cite{OmniH2O,ze2025twist}, we adopt a two-stage teacher–student learning paradigm to address the challenge of partial observability. In the first stage, an oracle teacher policy $\pi_{\mathrm{tea}}$ is trained with Proximal Policy Optimization (PPO)~\cite{ppo}, utilizing full state information. In the second stage, a student policy $\pi_{\mathrm{stu}}$ is trained via behavior cloning (BC) to imitate the teacher, relying only on observations available at deployment. The overall framework of \method{} is illustrated in Fig.~\ref{fig:framework}.

\subsection{Motion Dataset Curation}

To develop a versatile humanoid motion controller, we first construct a large-scale, high-quality human motion dataset as the training source. Starting from the publicly available AMASS Mocap dataset~\cite{amass} in SMPL format, we retarget human motions to the humanoid skeleton using a differentiable inverse kinematics (IK)-based method~\cite{ze2025gmr,Zakka_Mink_Python_inverse_2024}. This procedure formulates a differentiable optimization problem, ensuring end-effector trajectory alignment under the constraints of joint limits, yielding over 13,000 sequences.

The raw retargeted dataset, however, contains motions infeasible for the humanoid (e.g., stair climbing) and sequences with corrupted frames (e.g., severe penetrations or discontinuities~\cite{luo2024universal}). To curate the data, we train an oracle policy (see Section~\ref{sec:tsl}) on the full set and evaluate each sequence, filtering out invalid ones. The final dataset comprises 9,770 high-quality motions, totaling 30.41 hours of humanoid-compatible data.

\subsection{Policy Objective}

\subsubsection{Hybrid Tracking Targets}
To guide the policy toward both expressive motion style and accurate spatial alignment, we adopt a hybrid tracking objective that combines global root tracking with local keybody tracking~\cite{exbody2,liao2025beyondmimicmotiontrackingversatile}. 

For global root tracking, we denote the root's position and rotation in the world frame as $\bm{p}$ and $\bm{r}$. The policy seeks to minimize deviations from the reference root trajectory, ensuring overall spatial alignment across the motion.  

Specifically, we define the set of keybody parts as $\mathcal{K}$, including the head, hands, elbows, knees, and ankles. The local tracking targets, denoted as keybody positions $\bm{p}^\mathcal{K}$ and orientations $\bm{r}^\mathcal{K}$, are obtained by aligning the positions and orientations of these keybodies relative to the root between the current robot state and the reference motion. This encourages the policy to capture the local motion style demonstrated in the reference. 
By defining both global and local tracking targets in this way, the policy is guided to leverage precise reference information, balancing local motion fidelity and global spatial consistency.

\subsubsection{State Space Design}

At each timestep $t$, the teacher policy $\pi_{\mathrm{tea}}$ observes a state $ \bm s^{\mathrm{tea}}_t$ and a goal $\bm g^{\mathrm{tea}}_t$:
\begin{equation}
\bm s^{\mathrm{tea}}_t =
\Big[
\bm{q}_t, \dot{\bm{q}}_t, \bm{v}_{t}, \bm{\omega}_{t}, 
\bm{p}^\mathcal{K}_t, \bm{r}^\mathcal{K}_t, \Delta \bm{p}_t, \Delta \bm{r}_t, \bm{a}_{t-1}
\Big],
\end{equation}
\begin{equation}
\bm g^{\mathrm{tea}}_t =
\Big[
\bm{q}^{\mathrm{ref}}_{t+1:t+H}, \dot{\bm{q}}^{\mathrm{ref}}_{t+1:t+H}, \bm{p}^{\mathcal{K},\mathrm{ref}}_{t+1:t+H}, \bm{r}^{\mathcal{K},\mathrm{ref}}_{t+1:t+H}
\Big],
\end{equation}
where $\bm{q}_t \in \mathbb{R}^{23}$ and $\dot{\bm{q}}_t \in \mathbb{R}^{23}$ denote the joint positions and velocities, $\bm{v}_t \in \mathbb{R}^{3}$ and $\bm{\omega}_t \in \mathbb{R}^{3}$ are the root's linear and angular velocities, and $\bm{a}_{t-1} \in \mathbb{R}^{23}$ is the previous action.  
$\Delta \bm{p}_t \in \mathbb{R}^3$ and $\Delta \bm{r}_t \in \mathbb{R}^6$~\cite{zhou2019continuity} represent the root's position and rotation tracking errors.  
The goal $\bm{g}^{\mathrm{tea}}_t$ provides a preview of reference joint positions, velocities, and relative keybody poses over the future $H$ timesteps.

In real-world deployment, global root position and velocity are unavailable. Therefore, the student policy $\pi_{\mathrm{stu}}$ relies on proprioceptive states, augmented with a short history of past observations. Its state $ \bm s^{\mathrm{stu}}_t$ and goal $\bm g^{\mathrm{stu}}_t$ are defined as:
\begin{equation}
\bm s^{\mathrm{stu}}_t =
\Big[
\bm{q}_{t-K:t}, \dot{\bm{q}}_{t-K:t}, \bm{\omega}_{t-K:t}, \Delta \bm{r}_{t-K:t}, \bm{a}_{t-K:t-1}
\Big],
\end{equation}
\begin{equation}
\bm g^{\mathrm{stu}}_t =
\Big[
\bm{q}^{\mathrm{ref}}_{t+1:t+H}, \dot{\bm{q}}^{\mathrm{ref}}_{t+1:t+H}, \bm{r}^{\mathcal{K},\mathrm{ref}}_{t+1:t+H}
\Big],
\end{equation}
where $K$ is the number of past timesteps included in the student state, enabling the policy to leverage historical proprioceptive information to compensate for the partial observability. By defining goal states for both the teacher and the student, policies can plan actions conditioned on future motion, improving tracking accuracy and expressiveness.

\subsection{Policy Learning Framework}
\label{sec:tsl}
\subsubsection{Orthogonal Mixture of Experts}
Learning versatile motion skills can be considered as a multi-task RL problem, where the policy must imitate diverse motions simultaneously. A single MLP often struggles, as it mixes multiple motions into overlapping representations, hindering effective skill learning. For instance, AMASS dataset~\cite{amass} span categories such as walking, running, kicking, squatting, and dancing, each with distinct dynamics patterns. Collapsing them into a single shared representation often leads to instability and limited expressiveness~\cite{chen2025gmtgeneralmotiontracking}. To address this, we introduce an \emph{Orthogonal Mixture-of-Experts} (OMoE) architecture for the teacher policy. The OMoE module consists of multiple expert networks whose outputs are constrained to be orthogonal, along with a router network that dynamically selects experts based on the current state.

At each timestep $t$, the input to OMoE is the concatenation of the state and goal $( \bm s_t, \bm g_t)$. A set of $M$ experts $\{h_i\}_{i=1}^M$ maps the input to feature vectors:
\begin{equation}
U_t = [\bm{u}_1, \bm{u}_2, \dots, \bm{u}_M] \in \mathbb{R}^{d \times M}, \quad \bm{u}_i = h_i(\bm s_t, \bm g_t).
\end{equation}
To encourage diversity, we impose an orthogonality constraint on the output features of different experts:
\begin{equation}
U_t^\top U_t = I_M, 
\end{equation}
ensuring that each $\bm{u}_i$ represents an independent direction in feature space. Directly solving this constrained optimization is challenging, so we instead approximate it via the Gram–Schmidt (GS) process~\cite{leon2013gram,hendawy2024multitaskreinforcementlearningmixture}. GS maps $U_t$ into an orthogonal basis $V = \{\bm{v}_1, \dots, \bm{v}_M\}$ by sequentially removing projections onto previously obtained vectors:
\begin{equation}
\bm{v}_i = \bm{u}_i - \sum_{j=1}^{i-1} 
\frac{\langle \bm{v}_j, \bm{u}_i \rangle}{\langle \bm{v}_j, \bm{v}_j \rangle}\,\bm{v}_j, 
\quad i=1,\dots,M.
\end{equation}
This ensures that different experts provide mutually diverse representations. In practice, the differentiable nature of GS encourages the experts to learn diverse features, while normalization after projection further stabilizes training.

A router network computes the expert weights as 
$\bm\alpha = \mathrm{Router}(\bm s_t, \bm g_t)$
and assigns the resulting coefficients $\alpha_i$ to their orthogonalized features. The weighted combination is passed through a shared output head $f$, producing the teacher action $\bm a^\mathrm{tea}_t$:
\begin{equation}
\bm a^\mathrm{tea}_t = f \!\left(\sum_{i} \alpha_i \bm v_i \right).
\end{equation}

With the design of OMoE, the teacher policy decomposes diverse motion skills into orthogonal representations, and the router adaptively recombines them, allowing flexible composition of versatile skills.

\subsubsection{History Encoding and Student Distillation}
During teacher training, a short temporal history of the robot’s proprioceptive states is encoded by a convolutional network into a compact latent embedding. This embedding is trained to approximate privileged, simulation-only randomized physical parameters, such as motor strengths, friction coefficients, and base mass variations. In this way, the latent captures environment dynamics and physical variations that are unavailable during real-world deployment.

In the student training stage, this history encoder is frozen and applied to real-world history observations, producing latent embeddings that enhance robustness to system noise and parameter uncertainty. Leveraging this representation enables the student to more effectively imitate the teacher. The student policy is then distilled via DAgger~\cite{ross2011reduction}, minimizing the $\mathcal{L}_{2}$ loss between the student’s output actions $\bm a^\mathrm{stu}_t$ and the teacher’s actions via $ \| \bm a^\mathrm{tea}_t -\bm a^\mathrm{stu}_t\|_2^2$.

\subsection{Segment-level Tracking Reward}
Strict step-wise tracking of the reference trajectory often fails in long-horizon motion tracking due to error accumulation or infeasible reference states. For example, when the reference requires overly aggressive velocities or the robot is disturbed by external forces, one-to-one state matching may hinder balance and destabilize the policy. While global tracking emphasizes overall trajectory coverage and local tracking preserves short-term style, both objectives can be overly restrictive when the reference becomes difficult or impractical to follow.

To mitigate this, we utilize a \emph{segment-level tracking reward} that relaxes rigid step-wise tracking. As illustrated in Fig. \ref{fig:framework}, instead of enforcing alignment at a single timestep $t$, the agent is rewarded according to the minimum discrepancy between its state and the candidate reference states within a short temporal neighborhood.
For global tracking, the reward $r^{\text{global}}_t$ is defined as
\begin{equation}
r^{\text{global}}_t = \exp\!\Big(- \min_{\tau \in [0,H]} \; d_\text{global}(\bm{p}_t, \bm{p}^\mathrm{ref}_{t+\tau})\Big),
\end{equation}
where $H$ is the small future horizon and $d_\text{global}$ measures root position deviation. This allows the agent to catch up within a short future window, reducing sensitivity to transient errors.

Similarly, for local keybody tracking, we define
\begin{equation}
r^{\text{local}}_t = \exp\!\Big(- \min_{\tau \in [0,H]} \; d_\text{local}(\bm{p}^\mathcal{K}_t, \bm{p}^{\mathcal{K},\text{ref}}_{t+\tau})\Big),
\end{equation}
where $d_\text{local}$ measures deviations in local keybody positions. 

The segment-level design mitigates the limitations of strict step-wise matching by allowing the policy to align with the most feasible reference within a short horizon. This alleviates excessive penalties from temporary deviations or infeasible local targets, while jointly balancing local style preservation and global trajectory consistency, leading to more robust long-horizon tracking. The detailed reward terms and weights are summarized in Table~\ref{tab:reward}.

\begin{table}[h]
\centering
\caption{Reward terms and weights.}
\label{tab:reward}
\begin{tabular}{cccc}
\toprule
\textbf{Tracking terms} & \textbf{Weights} & \textbf{Regularization terms} & \textbf{Weights} \\
\midrule
Root position & $1.5$ & Action rate & $-0.1$ \\
Root velocity & $1.5$ & Feet slip & $-0.1$ \\
Root rotation & $1.5$ & Torque limits  & $-5.0$ \\
Key body position & $3.0$ & Joint limits & $-10$ \\
Key body velocity & $2.0$ & Joint velocities  & $-1\mathrm{e}{-4}$  \\
Key body rotation & $2.0$ & Joint accelerations  & $-3\mathrm{e}{-7}$\\
\bottomrule
\end{tabular}
\end{table}

\subsection{Sim2Real Transfer}

To improve policy robustness and facilitate effective sim-to-real transfer, we employ domain randomization (DR)~\cite{Peng_2018} during both teacher and student training. Additionally, we randomize default joint positions and perturb the robot root with additional linear and angular velocities. 

Furthermore, we adopt a noised reference state initialization (RSI) strategy~\cite{deepmimic,wang2025hilhybridimitationlearning}, injecting Gaussian noise into the root and joint states of the sampled motion states. This enhances the policy's resilience to imperfect intermediate states and environmental variations. The detailed configurations are provided in Table~\ref{tab:dr}.  

\begin{table}[h]
\centering
\caption{Randomization parameters.}
\begin{tabular}{cccc}
\toprule
\textbf{DR terms} & \textbf{Range} & \textbf{Noised RSI} & \textbf{Noise scale} \\
\midrule
Base Mass & $[-3, 3]$ & Joint position & $ 0.1$ \\
Friction & $[0.1, 1.6]$ &  Root position & $0.05$ \\
Motor Strength & $[0.9, 1.1]$ & Root velocity & $0.2$ \\
Default joint pos & $[-0.01, 0.01]$ & Root rotation & $0.1$ \\
Push Robot & $[-0.5, 0.5]$  & Root angular vel & $0.5$ \\
\bottomrule
\end{tabular}
\label{tab:dr}
\end{table}
\section{EXPERIMENTS}
In our experiments, we aim to answer the following key research questions: 
\begin{itemize}
\item \textbf{Q1.} Can \method{} outperforms baseline methods when trained on large-scale motion datasets?
\item \textbf{Q2.} Does OMoE enables effective learning and generalization of versatile motion skills?
\item \textbf{Q3.} Can the hybrid tracking objective augmented with a segment-level reward improve long-horizon motion performance?
\item \textbf{Q4.} How well does \method{} perform in real-world?
\end{itemize}

\subsection{Experiment Setup}

We evaluate the performance of \method{} in both simulation and real world. In simulation, we train policies on the curated training dataset and assess generalization on a test dataset, including the AMASS test dataset~\cite{luo2024universal}, the LAFAN1 dataset~\cite{liao2025beyondmimicmotiontrackingversatile}, and additional in-house Mocap motions. Each policy is trained in IsaacGym~\cite{isaac} simulator with 4,096 parallel environments, and for evaluation we sample 1,000 rollout trajectories per motion.

\subsubsection{Baselines}

We compare \method{} with two baselines, ExBody2~\cite{exbody2} and GMT~\cite{chen2025gmtgeneralmotiontracking}. For fairness, we re-implement both methods on our curated training dataset and conducted evaluations on the teacher policies, as the student policies are purely distilled from the teacher through behavior cloning.

\subsubsection{Metrics} 

Tracking performance is evaluated using the following metrics: Mean Per Keybody Position Error ($E_\mathrm{mpkpe}$, $mm$), Mean Per Joint Position Error ($E_\mathrm{mpjpe}$, $rad$), Mean Global Root Position Error ($E_\mathrm{pos}$, $mm$) and Mean Root Linear Velocity Error ($E_\mathrm{vel}$, $m/s$). For the test dataset, we report the success rate ($\%$), where a motion is considered successful if the humanoid completes the entire sequence.

\subsection{Main Results}
To address \textbf{Q1}, Table~\ref{tab:main-result} shows that \method{} outperforms ExBody2 and GMT across all metrics. It achieves notably better local tracking performance, while the inclusion of a global root objective effectively reduces root position errors. On the test dataset, although ExBody2 and GMT report relatively high success rates, they replace global position tracking with root velocity objectives, which amplifies global drift and leads to larger position errors. In contrast, \method{} substantially reduces both global and local tracking errors, demonstrating its accuracy and stability.

\begin{table*}[t]
\centering
\caption{Main results on training and test datasets. Results are reported as mean $\pm$ one standard deviation.}
\begin{tabular}{lcccc @{\hskip 2em} ccccc}
\toprule
& \multicolumn{4}{c}{\cellcolor{lightgreen}\textbf{Training Dataset}} 
& \multicolumn{5}{c}{\cellcolor{lightblue}\textbf{Test Dataset}} \\
\cmidrule(r){1-5} \cmidrule(r){6-10}
\textbf{Method} 
& $E_{\mathrm{mpkpe}} \downarrow$ 
& $E_{\mathrm{mpjpe}} \downarrow$ 
& $E_{\mathrm{pos}} \downarrow$ 
& $E_{\mathrm{vel}} \downarrow$ 
& $Succ \uparrow$ 
& $E_{\mathrm{mpkpe}} \downarrow$ 
& $E_{\mathrm{mpjpe}} \downarrow$ 
& $E_{\mathrm{pos}} \downarrow$ 
& $E_{\mathrm{vel}} \downarrow$ 
\\
\cmidrule(r){1-5} \cmidrule(r){6-10}
\rowcolor{gray!20}
\multicolumn{10}{l}{Baseline} \\
ExBody2 & 
{53.03}\textsubscript{$\pm$2.48}  & 
{0.684}\textsubscript{$\pm$0.020}  & 
{373.9}\textsubscript{$\pm$28.1}  & 
{0.256}\textsubscript{$\pm$0.020}  & 
{90.97}\textsubscript{$\pm$0.18}  & 
{62.78}\textsubscript{$\pm$0.805}  & 
{0.654}\textsubscript{$\pm$0.012}  & 
{468.2}\textsubscript{$\pm$40.2}  & 
{0.416}\textsubscript{$\pm$0.028} 
\\

GMT & 
{45.28}\textsubscript{$\pm$1.21}  & 
{0.526}\textsubscript{$\pm$0.011}  & 
{292.1}\textsubscript{$\pm$14.2}  & 
{0.187}\textsubscript{$\pm$0.006}  & 
{90.07}\textsubscript{$\pm$0.15}  & 
{58.89}\textsubscript{$\pm$0.932}  & 
{0.574}\textsubscript{$\pm$0.023}  & 
{397.1}\textsubscript{$\pm$28.7}  & 
{0.354}\textsubscript{$\pm$0.022} 
\\

\method{} (ours)    & 
\textbf{42.59}\textsubscript{$\pm$1.10}  & 
\textbf{0.499}\textsubscript{$\pm$0.009}  & 
\textbf{48.15}\textsubscript{$\pm$1.26}  &
\textbf{0.176}\textsubscript{$\pm$0.004}  & 
\textbf{92.68}\textsubscript{$\pm$0.20}  & 
\textbf{52.95}\textsubscript{$\pm$0.497}  & 
\textbf{0.568}\textsubscript{$\pm$0.003}  & 
\textbf{57.92}\textsubscript{$\pm$1.16}  & 
\textbf{0.216}\textsubscript{$\pm$0.002}  
\\
\midrule
\rowcolor{gray!20}
\multicolumn{10}{l}{Abaltions on OMoE} \\
     
\method{}-MLP &
{49.12}\textsubscript{$\pm$1.43}  & 
{0.583}\textsubscript{$\pm$0.020}  & 
{59.86}\textsubscript{$\pm$1.43}  & 
{0.199}\textsubscript{$\pm$0.010}  & 
{86.67}\textsubscript{$\pm$0.34}  & 
{56.73}\textsubscript{$\pm$0.811}  & 
{0.653}\textsubscript{$\pm$0.006}  & 
{67.21}\textsubscript{$\pm$3.92}  & 
{0.285}\textsubscript{$\pm$0.007}  
\\

\method{}-MoE    & 
{45.03}\textsubscript{$\pm$1.78}  & 
{0.558}\textsubscript{$\pm$0.014}  & 
{56.97}\textsubscript{$\pm$1.56}  & 
{0.188}\textsubscript{$\pm$0.007}  & 
{88.52}\textsubscript{$\pm$0.31}  & 
{54.86}\textsubscript{$\pm$0.517}  & 
{0.628}\textsubscript{$\pm$0.003}  & 
{59.90}\textsubscript{$\pm$1.25}  & 
{0.250}\textsubscript{$\pm$0.003}  
\\

\method{} (ours)    & 
\textbf{42.59}\textsubscript{$\pm$1.10}  & 
\textbf{0.499}\textsubscript{$\pm$0.009}  & 
\textbf{48.15}\textsubscript{$\pm$1.26}  &
\textbf{0.176}\textsubscript{$\pm$0.004}  & 
\textbf{92.68}\textsubscript{$\pm$0.20}  & 
\textbf{52.95}\textsubscript{$\pm$0.497}  & 
\textbf{0.568}\textsubscript{$\pm$0.003}  & 
\textbf{57.92}\textsubscript{$\pm$1.16}  & 
\textbf{0.216}\textsubscript{$\pm$0.002}  
\\
\midrule
\rowcolor{gray!20}
\multicolumn{10}{l}{Abaltions on Segment-level Tracking} \\
\method{}-Global & 
{50.28}\textsubscript{$\pm$1.55}  & 
{0.562}\textsubscript{$\pm$0.022}  & 
\textbf{44.09}\textsubscript{$\pm$2.12}  & 
{0.180}\textsubscript{$\pm$0.010}  & 
{60.52}\textsubscript{$\pm$0.24}  & 
{69.57}\textsubscript{$\pm$0.969}  & 
{0.576}\textsubscript{$\pm$0.005}  & 
{60.53}\textsubscript{$\pm$1.12}  & 
{0.243}\textsubscript{$\pm$0.005}  
\\

\method{}-Step-wise & 
{43.42}\textsubscript{$\pm$1.01}  & 
{0.512}\textsubscript{$\pm$0.015}  & 
{51.15}\textsubscript{$\pm$2.43}  & 
{0.182}\textsubscript{$\pm$0.010}  & 
{84.68}\textsubscript{$\pm$0.27}  & 
\textbf{50.63}\textsubscript{$\pm$0.312}  & 
{0.577}\textsubscript{$\pm$0.004}  & 
{63.92}\textsubscript{$\pm$1.00}  & 
{0.266}\textsubscript{$\pm$0.002}  
\\

\method{} (ours)& 
\textbf{42.59}\textsubscript{$\pm$1.10}  & 
\textbf{0.499}\textsubscript{$\pm$0.009}  & 
{48.15}\textsubscript{$\pm$1.26}  &
\textbf{0.176}\textsubscript{$\pm$0.004}  & 
\textbf{92.68}\textsubscript{$\pm$0.20}  & 
{52.95}\textsubscript{$\pm$0.497}  & 
\textbf{0.568}\textsubscript{$\pm$0.003}  & 
\textbf{57.92}\textsubscript{$\pm$1.16}  & 
\textbf{0.216}\textsubscript{$\pm$0.002}  
\\

\bottomrule
\end{tabular}
\label{tab:main-result}
\end{table*}

\subsection{Ablation Studies}

\subsubsection{Ablations on OMoE}
\begin{figure}
    \centering
    \includegraphics[width=\linewidth]{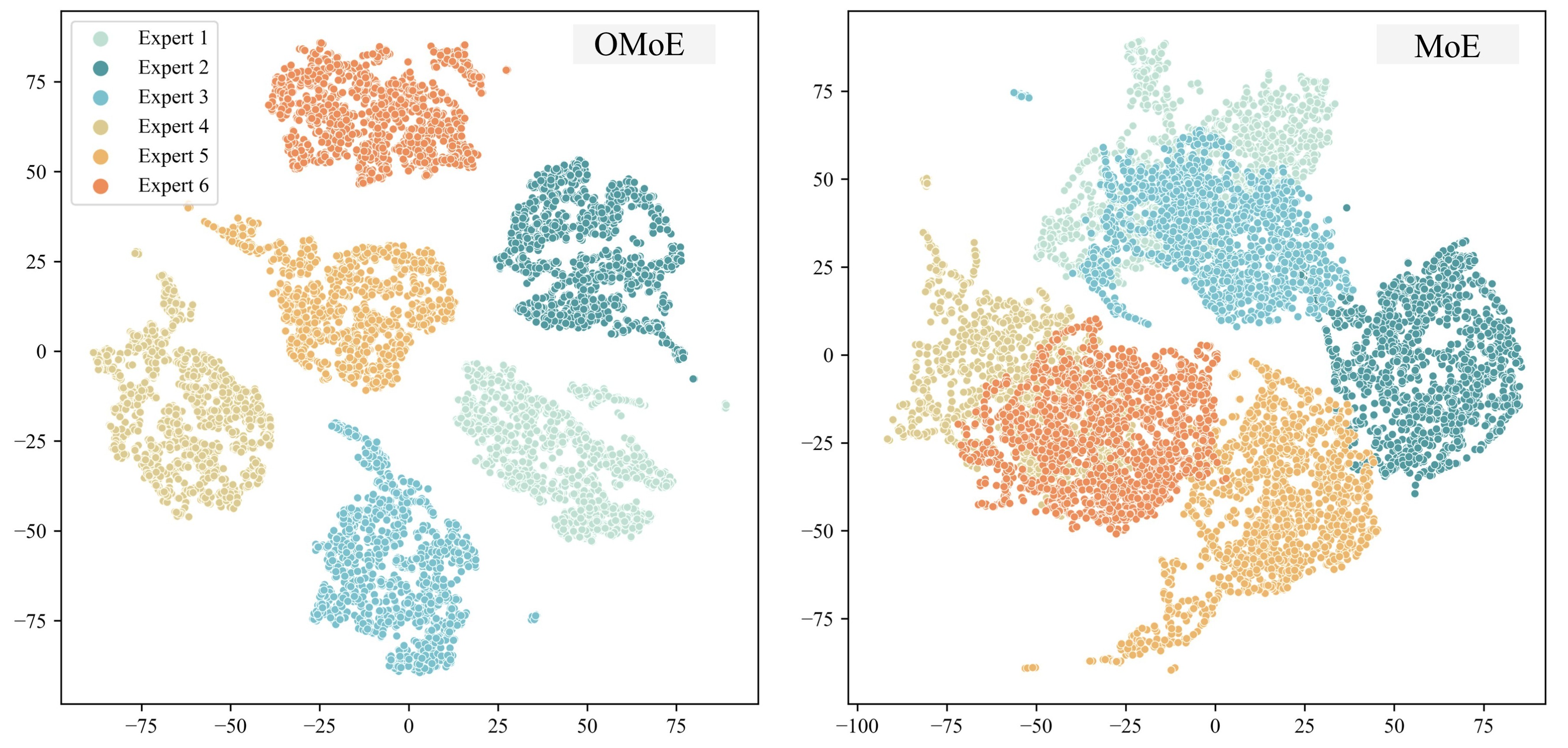}
    \caption{\textbf{t-SNE visualization of experts' output features.}  Standard MoE exhibits substantial overlap across experts, while OMoE produces more diverse and specialized skill subspaces.
    }
    \label{fig:tsne}
\end{figure}

To investigate \textbf{Q2}, we conduct ablation studies by comparing OMoE with a standard soft MoE~\cite{puigcerver2024sparsesoftmixturesexperts} and an MLP baseline of the same parameter size. As shown in Table~\ref{tab:main-result}, OMoE consistently achieves the lowest tracking errors, followed by MoE, while MLP performs the worst. This demonstrates that introducing orthogonalized experts leads to more effective motion tracking and better generalization.
In addition to quantitative metrics, we further analyze the learned representations from the experts. Specifically, we visualize the expert output features using t-SNE, as shown in Fig.~\ref{fig:tsne}. For the standard MoE, these features exhibit substantial overlap, indicating redundancy and limited differentiation among experts. In contrast, OMoE structure produces more dispersed and well-separated features, suggesting that the orthogonalization constraint encourages experts to capture distinct skill subspaces, thereby improving diversity and representation efficiency.

\begin{figure}
    \centering
    \includegraphics[width=\linewidth]{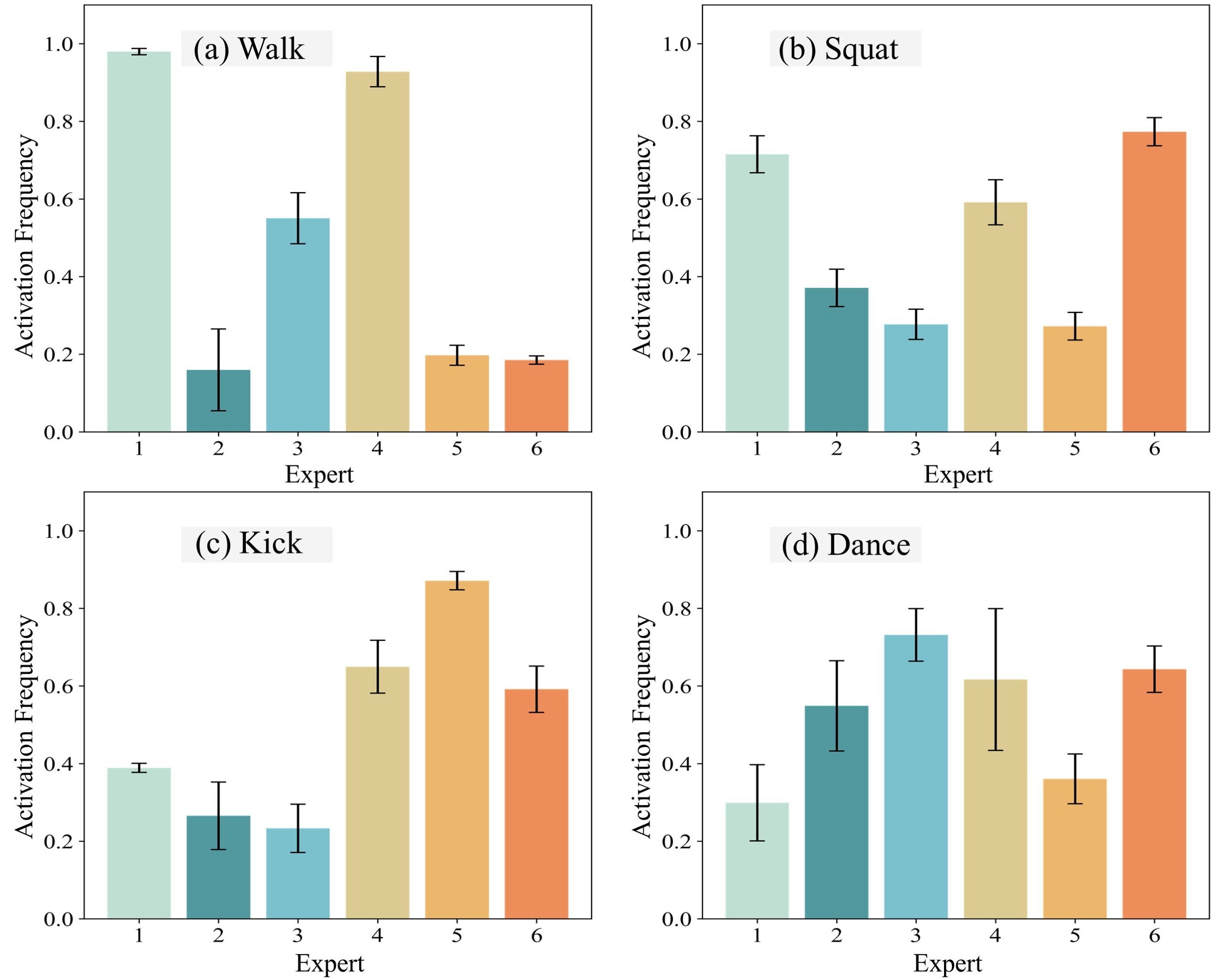}
    \caption{\textbf{Expert activation frequencies across representative motion categories.}  The OMoE architecture effectively decomposes skills and adapts expert usage to motion diversity.
    }
    \label{fig:weights}
\end{figure}

To further probe this specialization, we examine expert activation frequencies across four representative motion categories from AMASS: walking, squatting, kicking, and dancing. As shown in Fig.~\ref{fig:weights}, walking mainly activates a small subset of experts, reflecting its repetitive structure, while dancing exhibits a more evenly distributed activation pattern due to its higher variability. Squatting and kicking fall in between, each with distinct but less uniform patterns. These results confirm that \method{} adapts expert allocation to motion complexity, achieving both efficient specialization and flexible skill composition.

\subsubsection{Ablations on Segment-level Reward}

To answer \textbf{Q3}, we conduct an ablation study comparing \method{} with two other tracking strategies: (i) VMS-Global, where all keybodies are aligned strictly to their global targets~\cite{xie2025kungfubotphysicsbasedhumanoidwholebody,he2025asapaligningsimulationrealworld}, and (ii) VMS-Step-wise, which uses the same hybrid tracking objective as \method{} but enforces strict step-wise tracking target at each timestep. As shown in Table~\ref{tab:main-result}, while pure global tracking achieves the lowest root error on the training set, it generalizes poorly, exhibiting high errors and low success rates on the test set due to error accumulation and sensitivity to imperfect reference motions. Step-wise tracking achieves more stable performance but still underperforms our method across most metrics.

\begin{figure}
    \centering
    \includegraphics[width=\linewidth]{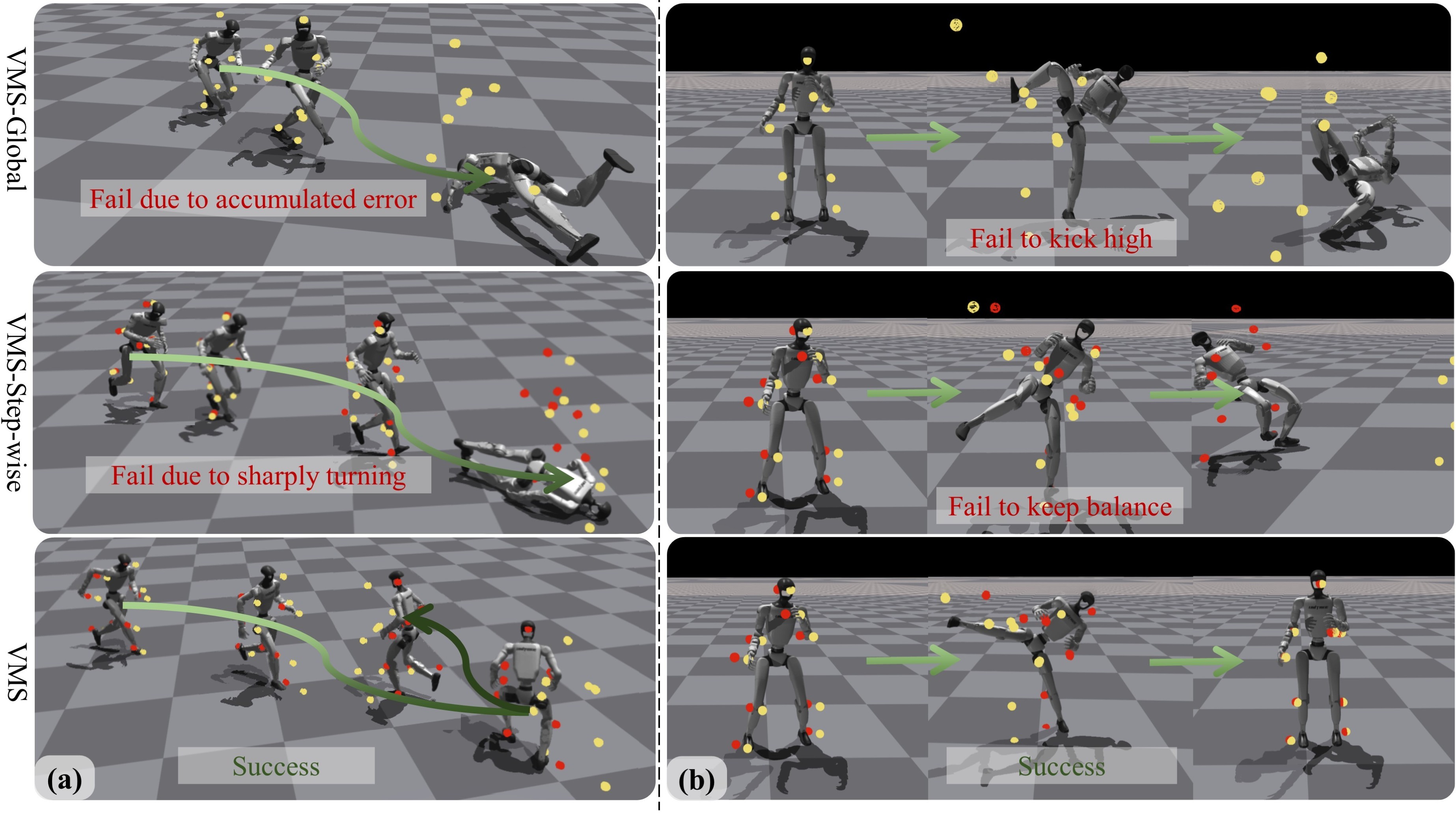}
    \caption{\textbf{Visualizations of example motions.} (a) for a running-to-turn motion, \method{} achieves a smooth transition, (b) for a side kick, it maintains balance and completes the motion stably.
    }
    \label{fig:Example_motion}
\end{figure}

\begin{figure}
    \centering
    \includegraphics[width=0.9\linewidth]{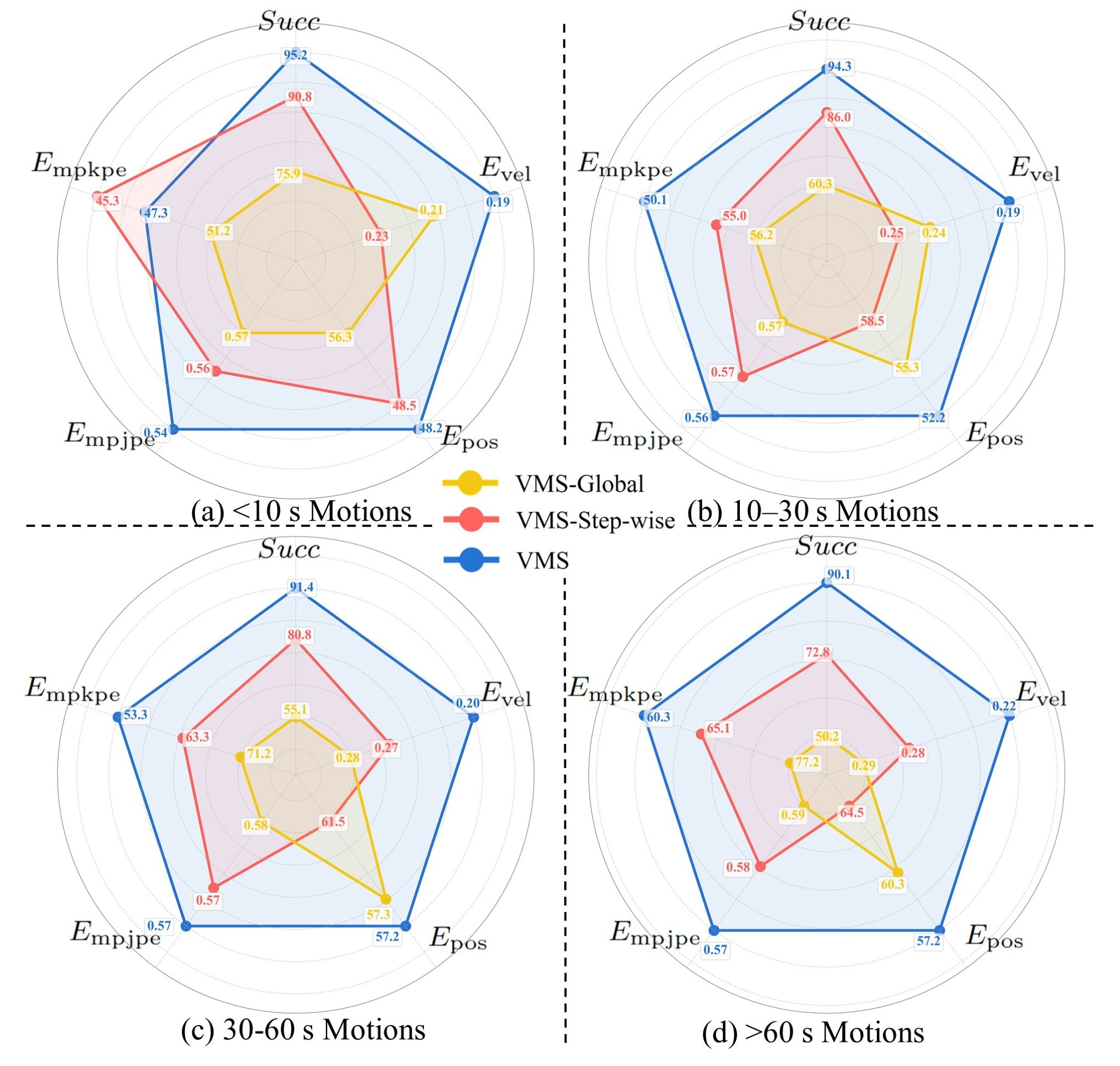}
    \caption{\textbf{ Tracking performance across motion durations.} Our segment-level tracking achieves consistently lower errors and higher robustness compared to global and step-wise tracking.
    }
    \label{fig:Error_motion_len}
\end{figure}

To further analyze, we visualize two representative motions in Fig.~\ref{fig:Example_motion}, with yellow points denoting global target keybody positions and red points representing local tracking targets. In example (a), a fast run followed by a sharp turn, the global tracker quickly fails as accumulated errors prevent it from reaching the target, while the step-wise tracker rigidly adheres to the target path and collapses during the turn. In contrast, our segment-level reward leverages a short future window, allowing smoother transitions and preserving motion style. In example (b), a side kick motion, the global tracker overemphasizes height and falls, while the step-wise tracker completes the kick but loses balance afterward. Our method succeeds by slightly compromising on kick height to maintain overall stability. In Fig.~\ref{fig:Error_motion_len}, test motions are grouped by duration. While both compared strategies degrade rapidly as motion duration increases, our method remains robust, showing strong performance on long-horizon motions.

\subsection{Real-World Experiments}
For \textbf{Q4}, Fig.~\ref{fig:cover} and the supplementary videos show our robot executing diverse real-world skills, including: (1) locomotion styles such as walking, marching, and running; (2) athletic movements like racket swings and ball throws; (3) expressive dances (e.g., Charleston); (4) highly dynamic actions such as punches, side kicks, and jumping kicks; and (5) long-horizon composite skills involving martial arts, such as Tai Chi and Shaolin Kungfu. These results demonstrate \method{}’s versatility as a general-purpose humanoid controller.

\subsection{Downstream Tasks}

\subsubsection{Text-to-motion Generation}
\begin{figure}
    \centering
    \includegraphics[width=\linewidth]{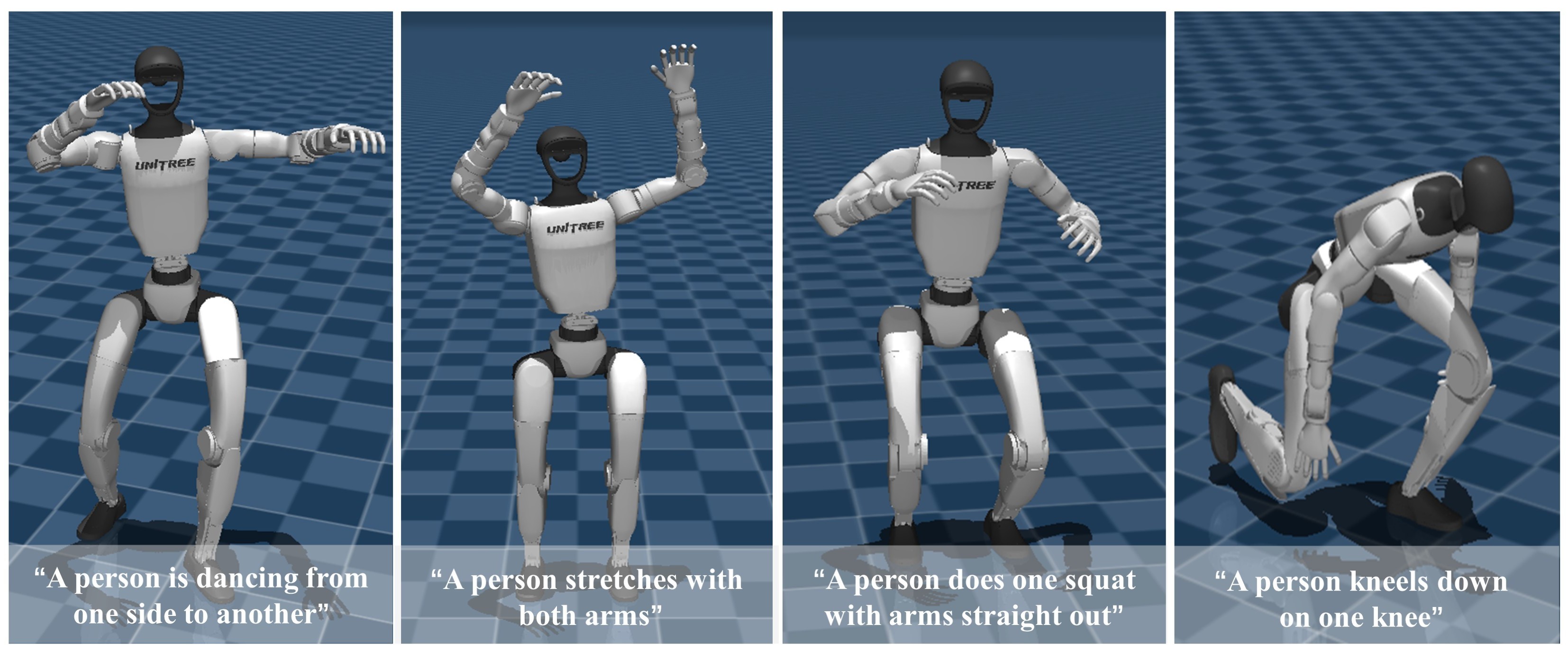}
    \caption{\textbf{Text2motion generation results.} \method{} reproduces diverse motions such as dancing, stretching, squatting, and kneeling from text descriptions, demonstrating strong zero-shot generalization.
    }
    \label{fig:text2motion}
\end{figure}

To further assess the generalization ability of \method{}, we conduct experiments on text-to-motion generation. The MLD model~\cite{Chen_2023_CVPR} is used to generate reference motions from language descriptions, and we evaluate whether the policy can follow them. As shown in Fig.~\ref{fig:text2motion}, \method{} successfully follows diverse text-generated motion instructions, highlighting its potential to serve as a universal low-level controller for higher-level planners.
\subsubsection{Finetuning on Extreme Motions}
\begin{figure}
    \centering
    \includegraphics[width=\linewidth]{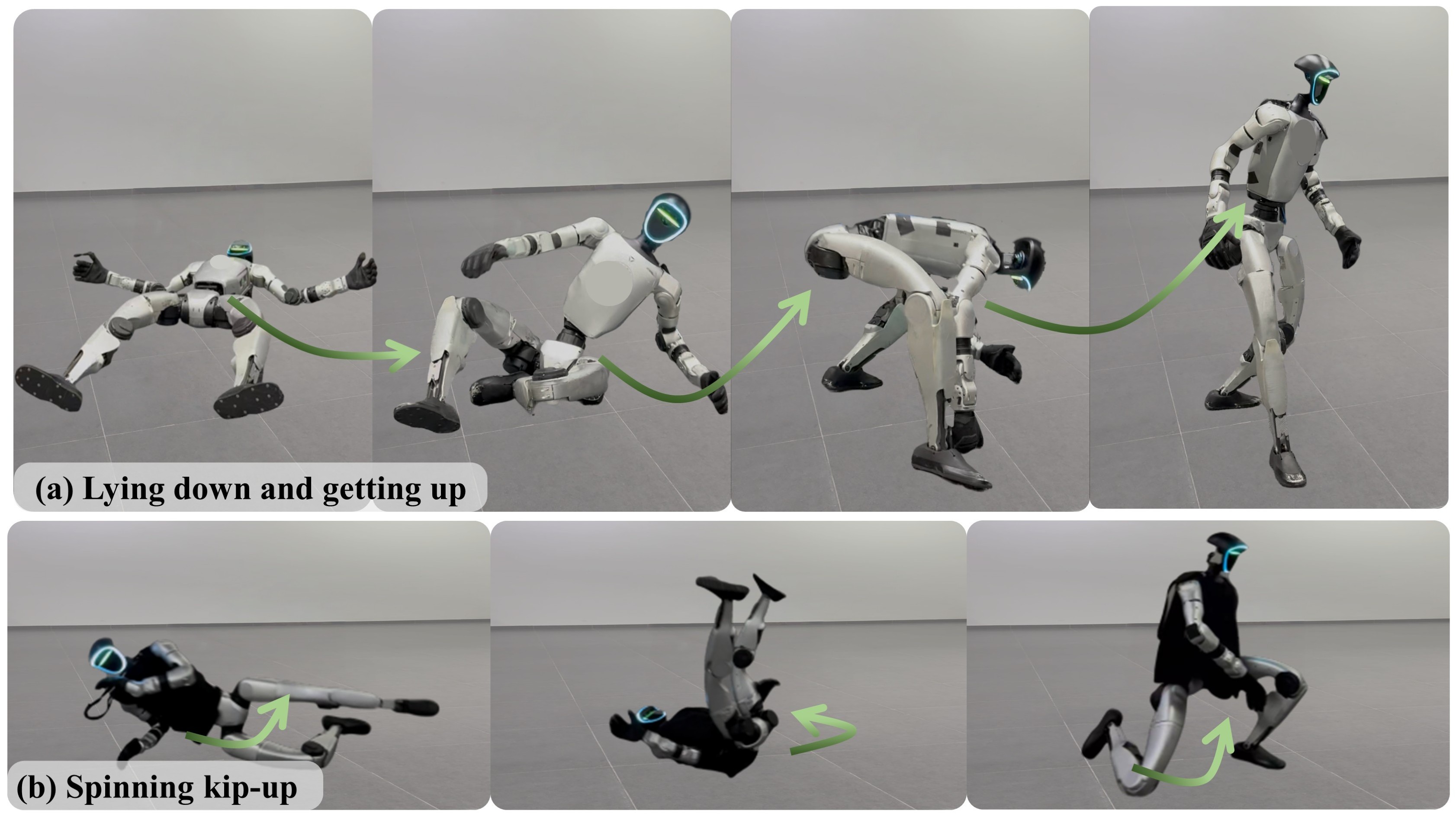}
    \caption{\textbf{Fintuning results.} \method{} adapting to OOD motions: (a) lying down and getting up, and (b) a highly dynamic spinning kip-up. 
    }
    \label{fig:fintune}
\end{figure}

Furthermore, \method{} also adapts well to challenging out-of-distribution (OOD) motions. Skills that demand collision ignoring (e.g., lying down and getting up) or highly acrobatic movements (e.g., spinning kip-up) can both be realized with minimal fine-tuning. In Fig.~\ref{fig:fintune}, the results highlight \method{}’s practicality to edge cases and downstream applications.
\section{Conclusion}

In this work, we present \method{}, a universal framework that enables humanoid robots to learn versatile motion skills. By combining an OMoE architecture for skill decomposition with a hybrid tracking objective and a segment-level reward, our approach achieves dynamic, human-like motions while maintaining stability over long-horizon sequences. Extensive experiments demonstrate that \method{} outperforms strong baselines, achieving lower tracking errors and higher success rates on unseen motions. Real-world deployments demonstrate that \method{} can perform various dynamic motions, including minute-level sequences. The downstream tasks highlight the potential of \method{} to serve as a foundation for general humanoid control.

However, \method{} still has limitations. First, it lacks visual perception, limiting its ability to understand complex scenes. Second, it heavily relies on large-scale Mocap datasets, which may contain uneven coverage of certain skills, potentially affecting generalization. Addressing these limitations is an important direction for future work.









\bibliographystyle{IEEEtran}
\bibliography{bibtex}

\end{document}